%% file: main.tex
\documentclass[11pt,a4paper]{article}
\usepackage[hyperref]{emnlp-ijcnlp-2019}
\usepackage{times}
\usepackage{latexsym}
\usepackage{graphicx}
\usepackage{amsmath}
\usepackage{amssymb}
\usepackage{multirow}
\usepackage{enumitem}

\aclfinalcopy 

\title{Uncover the Ground-Truth Relations in Distant Supervision: A Neural Expectation-Maximization Framework}

\author{Junfan Chen \\
  BDBC and SKLSDE\\
  Beihang University, China \\
  {\tt\small chenjf@act.buaa.edu.cn} \\
  \And
  Richong Zhang\thanks{$\ \ $Corresponding author}\\
  BDBC and SKLSDE\\
  Beihang University, China \\
  {\tt\small zhangrc@act.buaa.edu.cn} \\
  \And
Yongyi Mao \\
School of EECS\\
  University of Ottawa, Canada \\
  {\tt\small ymao@uottawa.ca} \\
  \AND
  Hongyu Guo \\
  Digital Technologies Research Center\\
  National Research Council Canada, Canada \\
  {\tt\small 	hongyu.guo@nrc-cnrc.gc.ca} \\
  \And 
  Jie Xu \\
  School of Computing\\
  University of Leeds, United Kingdom \\
  {\tt\small j.xu@leeds.ac.uk} \\
}  

\date{}

\begin{document}
\maketitle
\input{abs}
\input{intro} 
\input{problem} 
\input{related} 
\input{methodNew} 
\input{exp} 
\input{conclusion} 
\section*{Acknowledgment}
This work is supported partly by China 973 program (No. 2015CB358700), by the National Natural Science Foundation of China (No. 61772059, 61421003), by the Beijing Advanced Innovation Center for Big Data and Brain Computing (BDBC), by State Key Laboratory of Software Development Environment (No. SKLSDE-2018ZX-17) and by the Fundamental Research Funds for the Central Universities.
\bibliography{emnlp-ijcnlp-2019}
\bibliographystyle{acl_natbib}

\end{document}

%% file: abs.tex
\begin{abstract}

Distant supervision for relation extraction enables one to effectively 
acquire structured relations  out of 
 very large text corpora  with less human efforts. Nevertheless, most of the prior-art models for such tasks assume that the given text can be noisy, but their corresponding labels are clean. Such unrealistic assumption is contradictory with the fact that the  given labels are often noisy as well, thus leading to significant performance degradation of those models on  real-world data. 
To cope with this challenge, we propose a novel label-denoising framework that combines neural network with probabilistic modelling, which naturally takes into account the noisy labels during learning. 
We  empirically demonstrate that our approach significantly improves the current art  in uncovering the ground-truth relation labels. 
\end{abstract}

%% file: intro.tex
\section{Introduction}
Relation extraction aims at automatically 
extracting semantic relationships from a piece of text. 
Consider the sentence ``Larry Page, the chief executive officer of Alphabet, Google's parent company, was born in East Lansing, Michigan.''. The  knowledge triple ({\em Larry\_Page}, {\em employed\_by}, {\em Google}) can be extracted. 
Despite various efforts in building relation  extraction models~\cite{Zelenko:02, Zhou:05, Razvan:05, Zeng:14, Santos:15, Yang:16, Liu:15, Xu:15, Miwa:16, GaborBSQZC:18}, the difficulty of obtaining  abundant training data with labelled relations remains a challenge, and thus motivates the development of {\em Distant Supervision}  relation extraction~\cite{Mintz:09, Riedel:10, Zeng:15, Lin:16, Ji:17, Zeng:18, Feng:18, Wang:18a, Wang:18b, Hoffmann:11, Surdeanu:12, Jiang:16, Ye:17, Su:18, Qu:18}. 
Distant supervision (DS) relation extract methods collect a large dataset with ``distant'' supervision signal and learn a relation predictor from such data. In detail, each example in the dataset contains a collection, or {\em bag}, of sentences all involving the same pair of entities extracted from some corpus (e.g.,  news reports). 
Although such a dataset is expected to be very noisy, one hopes that when the dataset is large enough, useful correspondence between the semantics of a sentence and the relation label it implies still reinforces and manifests itself.
Despite their capability of learning from large scale data, we show in this paper that these DS relation extraction strategies fail to  adequately model the characteristic of the noise in the data. Specifically, most of the works fail to recognize that the labels can be noisy in a bag and directly use bag labels as training targets. 

 This aforementioned observation has inspired us to study a more realistic setting for DS relation extraction. That is, we treat the bag labels as the noisy observations of the ground-truth labels. 
To that end, we develop a novel framework that jointly models the semantics representation of the bag,  the latent  ground truth labels,  and the noisy observed labels. The framework, probabilistic in nature,  allows any neural network, that encodes the bag semantics,  to nest within. We show that the well-known Expectation-Maximization (EM) algorithm can be applied to the end-to-end learning of the models
built with this framework. As such, we term the framework  {\em Neural Expectation-Maximization}, or the {\em nEM}.  

Since our approach deviates from the conventional models and regards bag labels not as the ground truth, bags with the real ground-truth labels are required for evaluating our model. To that end, we manually re-annotate a fraction of the testing bags in a standard DS dataset with their real ground-truth labels. We then perform extensive experiments and demonstrate that the proposed nEM framework improves the current state-of-the-art models in uncovering the ground truth relations. 
\begin{figure}
\scalebox{1.0}{
\begin{tabular}{cc}
	 \hspace{-.1cm}
     \includegraphics[width=.43\columnwidth]{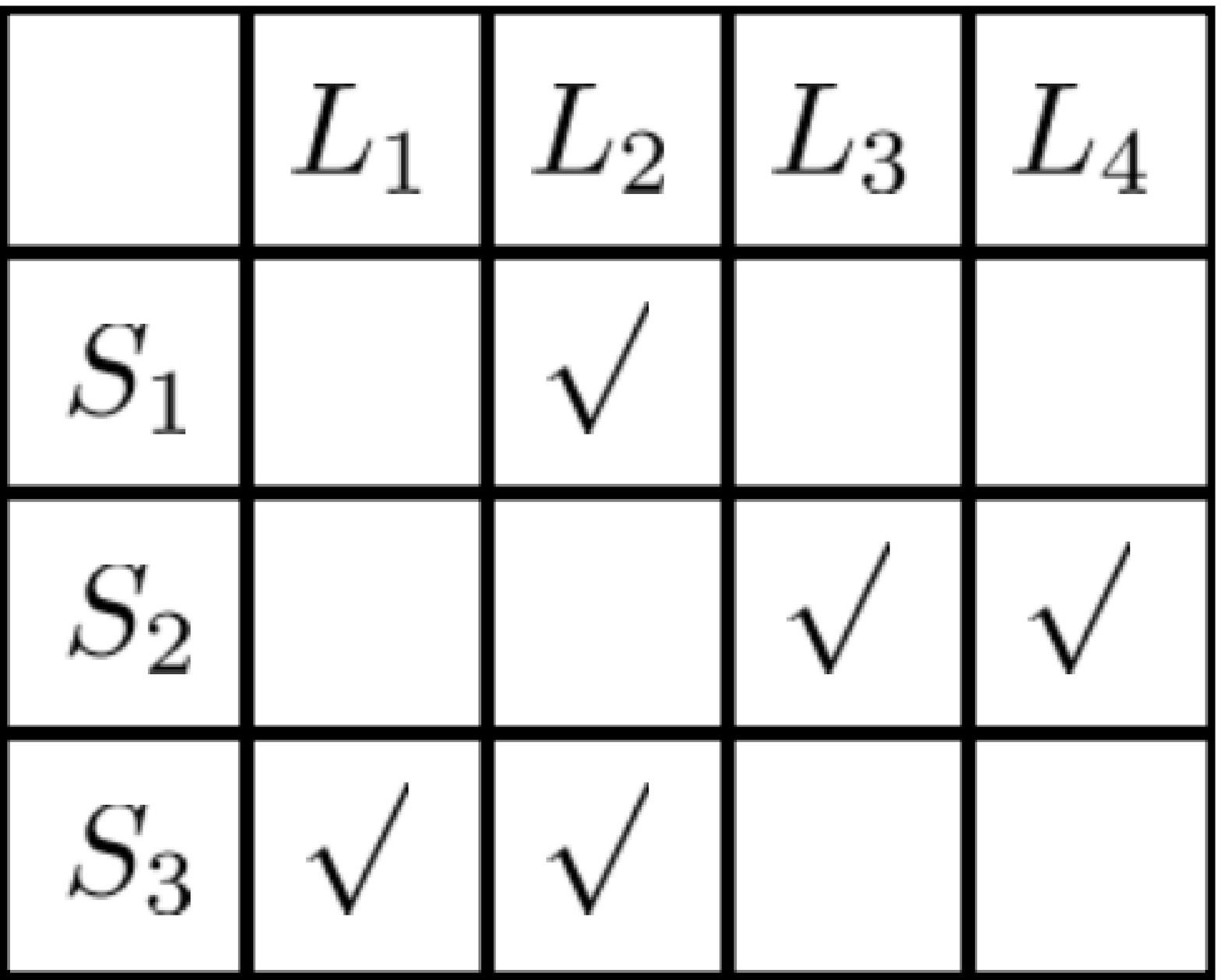}
     & 
     \hspace{-.1cm}
     \includegraphics[width=.43\columnwidth]{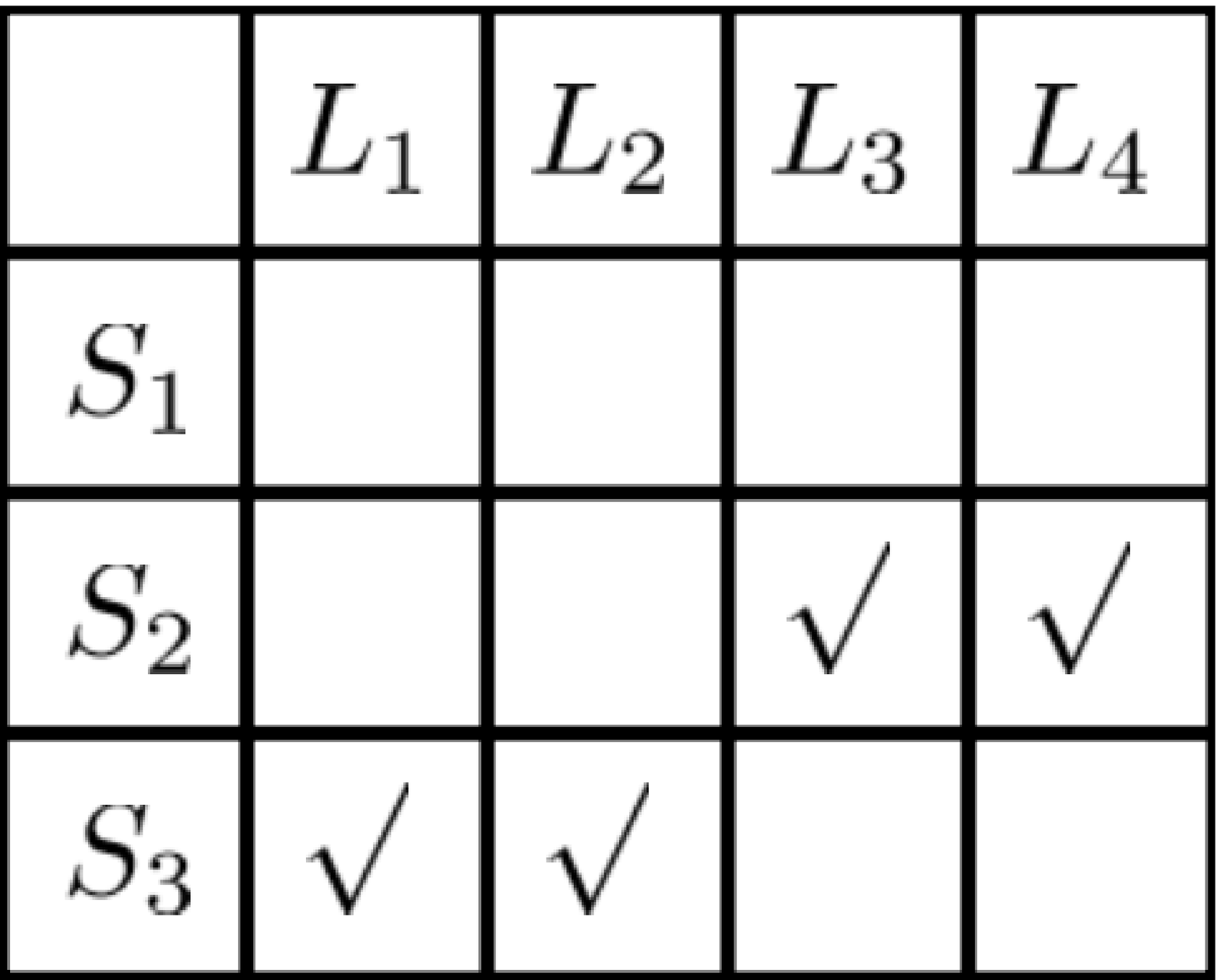} \\
     \hspace{-.1cm}
     (a) CSCL &(b) NSCL \\
      \includegraphics[width=.43\columnwidth]{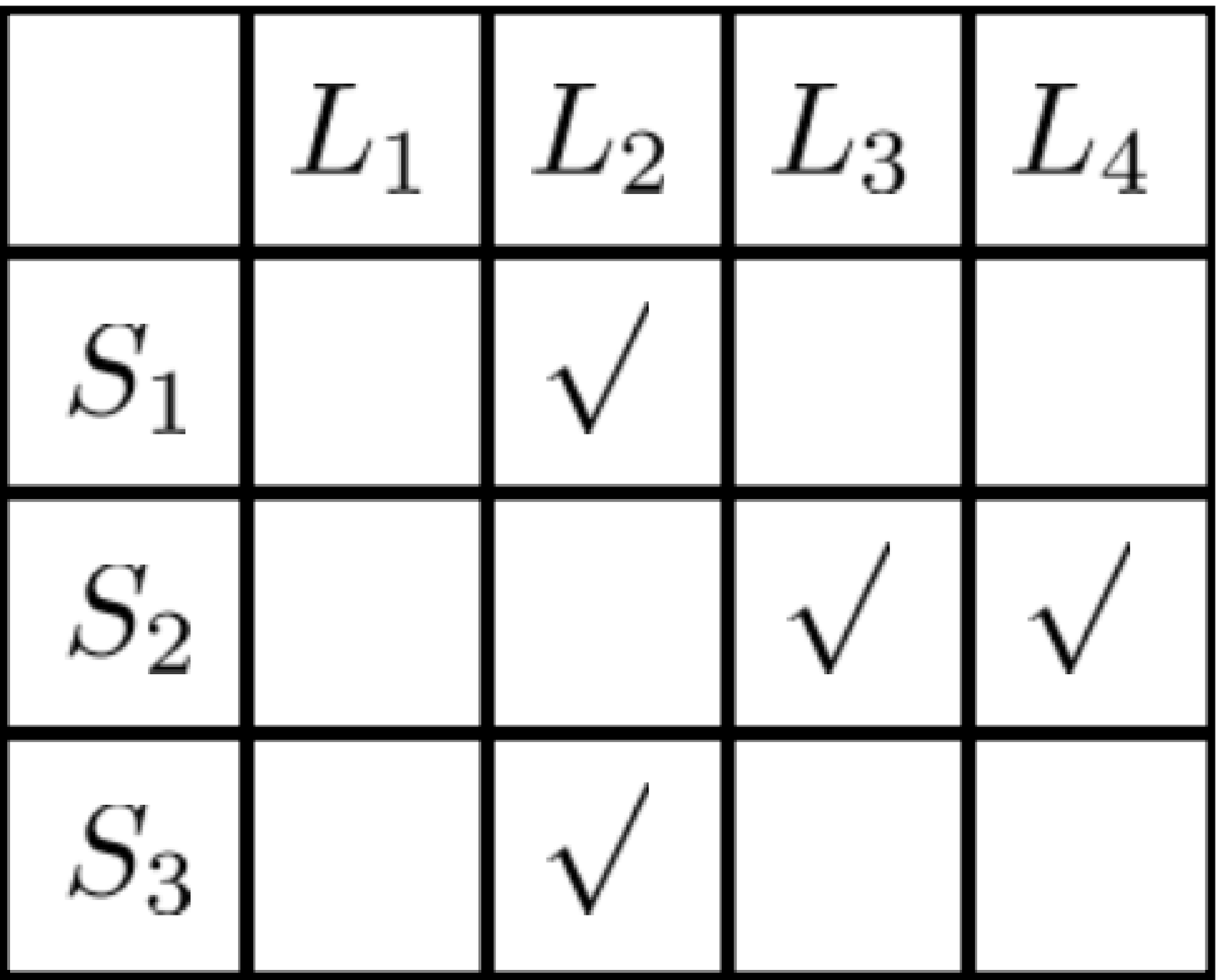}
     &
     \hspace{-.1cm}
     \includegraphics[width=.43\columnwidth]{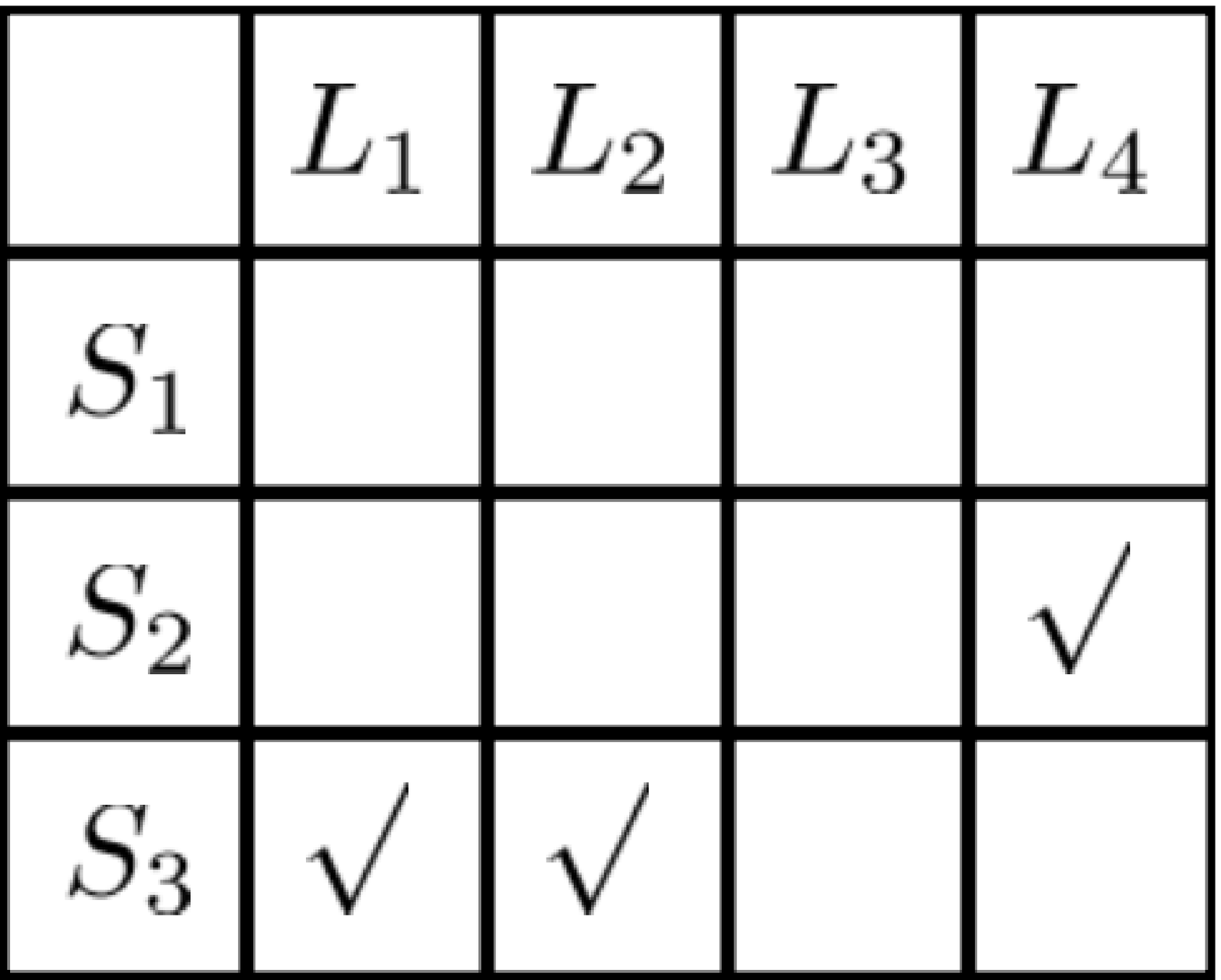}  \\
     (c) CSNL&(d) NSNL\\
\end{tabular}
}
\caption{Assumptions on the cleanness of DS dataset. In each matrix, rows correspond to the sentences in a bag, and columns correspond to the labels assigned to the bag. A check mark on $(S_i, Lj)$ indicates that the label $L_j$ is supported by sentence $S_i$. \label{fig:assumptions}}
\vspace{-.3cm}
\end{figure} 
To the best of our knowledge, this work is the first that combines a neural network model with EM training under the ``noisy-sentence noisy-label'' assumption.
The re-annotated testing dataset~\footnote{Will be released upon the acceptance of the paper}, containing the ground-truth relation labels, would also benefit the research community.

%% file: problem.tex
\section{Problem Statement}

Let  set ${\mathcal R}$ contain all relation labels of interest. Specifically, each label $r$ in ${\mathcal R}$ corresponds to a candidate relation in which any considered pair $(e, e')$ of entities may participate. Since ${\mathcal R}$ cannot contain all relations that are implied in a corpus, we include in ${\mathcal R}$ an additional relation label ``NA'', which refers to any relation that cannot be regarded as the other relations in ${\mathcal R}$. 

Any subset of ${\mathcal R}$ 
will be written as a $\{0, 1\}$-valued vector of length $|{\mathcal R}|$, for example, $z$, where each element of the vector $z$ corresponds to a label $r\in {\mathcal R}$. 
Specifically, if and only if label $r$ is contained in the subset, its corresponding element $z[r]$ of $z$ equals $1$. 
If two entities $e$ and $e'$ participate in a relation $r$, we say that the triple $(e, r, e')$ is factual.  
Let ${\mathcal B}$ be a finite set, in which  each $b\in {\mathcal B}$ is a pair $(e, e')$ of entities. Each $b = (e, e') \in {\mathcal B}$ serves as the index for a {\em bag}  $x_b$ of sentences.

The objective of DS is to use a large but noisy training set to learn a predictor that 
predicts the relations involving two arbitrary (possibly unseen) entities;
the predictor takes as input a bag of sentences each containing the two entities of interest, and hopefully outputs the set of all relations in which the two entities participate. 

%% file: related.tex
\section{Prior Art and Related Works} \label{related}
Relation extraction is an important task in natural language processing. Many approaches with supervised methods have been proposed to complete this task. These works, such as ~\cite{Zelenko:02, Zhou:05, Razvan:05}, although achieving 
good performance, rely on carefully selected features and well labelled dataset. 
Recently, neural network models, have been used in ~\cite{Zeng:14, Santos:15, Yang:16, Liu:15, Xu:15, Miwa:16} for supervised relation extraction. These models avoid  feature engineering and are shown to improve upon previous models.  But having a large number of parameters to estimate, these models rely heavily on costly human-labeled data.

Distant supervision was proposed in ~\cite{Mintz:09} to automatically generate large dataset through aligning the given knowledge base to text corpus. However, such dataset can be quite noisy. To articulate the nature of noise in DS dataset, a sentence is said to be {\em noisy} if it supports no relation labels of the bag, and a label of the bag is said to be {\em noisy} if it is not supported by any sentence in the bag. A sentence or label that is not noisy will be called {\em clean}. The cleanness of a training example may obey the following four assumptions, for each of which an example is given In Figure \ref{fig:assumptions}. 

\begin{itemize}[leftmargin=*, itemsep=-0.1cm]
\item 
\textbf{Clean-Sentence Clean-Label (CSCL)}: All sentences and all labels are clean (Figure \ref{fig:assumptions}(a)).

\item 
\textbf{Noisy-Sentence Clean-Label (NSCL)}: Some sentences may be noisy but all labels are clean (Figure \ref{fig:assumptions}(b)). Note that CSCL is 
a special case of NSCL. 
\item
\textbf{Clean-Sentence Noisy-Label (CSNL)}: All sentences are clean but some labels may be noisy(Figure \ref{fig:assumptions}(c)). Note that CSNL includes 
CSCL as a special case.
\item
\textbf{Noisy-Sentence Noisy-Label (NSNL)}: Some sentences may be noisy and some labels may also be noisy (Figure \ref{fig:assumptions}(d)). 
\end{itemize}
Obviously, CSCL, NSCL, CSNL are all special cases of NSNL. Thus
NSNL is the most general among all these assumptions. 

The author of ~\cite{Mintz:09} creates a model under the CSCL assumption, which is however pointed out a too strong assumption~\cite{Riedel:10}.
To alleviate this issue, 
many studies adopt NSCL assumption. Some of them, including ~\cite{Riedel:10, Zeng:15, Lin:16, Ji:17, Zeng:18, Feng:18, Wang:18a, Wang:18b}, formulate the task as a multi-instance learning problem where only one label is allowed for each bag. These works take sentence denoising through selecting the max-scored sentence~\cite{Riedel:10, Zeng:15, Zeng:18}, applying sentence selection with soft attention~\cite{Lin:16, Ji:17}, performing sentence level prediction as well as filtering noisy bags~\cite{Feng:18} and redistributing the noisy sentences into negative bags~\cite{Wang:18a, Wang:18b}. Other studies complete this task with multi-instance multi-label learning~\cite{Hoffmann:11, Surdeanu:12, Jiang:16, Ye:17, Su:18}, which allow overlapping relations in a bag. Despite the demonstrated successes, these models ignore the fact that relation labels can be noisy and ``noisy'' sentences that indeed point to factual relations may also be ignored. 
Two recent approaches~\cite{Liu:17,Luo:17} using the NSNL assumption have also been introduced, but these methods are evaluated based on the assumption that the evaluation labels are clean. 

\input{relation2NEMandDS.tex}


%% file: relation2NEMandDS.tex
We note that this paper is not the first work that combines neural networks with EM training. Very recently, a model also known as Neural Expectation-Maximization or NEM~\cite{Greff:17} has been presented to learn latent representations for clusters of objects (e.g., images) under a complete unsupervised setting. The NEM model is not directly applicable to our problem setting which deals with noisy supervision signals from categorical relation labels.  Nonetheless, given the existence of the acronym NEM, we choose to abbreviate our Neural Expecation-Maximization model as the nEM model.

%% file: methodNew.tex
\section{The nEM Framework}
We first introduce the nEM architecture and its learning strategy, and then present  approaches  to encode a bag of sentences (i.e., the  Bag Encoding Models) needed by the framework. 
\subsection{The nEM Architecture}
Let random variable $X$ denote a random bag and random variable $Z$ denote the label set assigned to $X$.  Under the NSNL assumption, $Z$, or some labels within,  may not be clean for $X$. We introduce another latent random variable $Y$, taking values as a subset of ${\mathcal R}$, indicating the set of ground-truth (namely, clean) labels for $X$. We will write $Y$ again as an $|{\mathcal R}|$-dimensional $\{0, 1\}$-valued vector.  
From here on, we will adopt the convention that a random variable will be written using a capitalized letter, and the value it takes will be written using the corresponding lower-cased letter. 

A key modelling assumption in nEM is that random variables $X$, $Y$ and $Z$ form a Markov chain
$X \rightarrow Y \rightarrow Z$. Specifically, the dependency of noisy labels $Z$ on the bag $X$ is modelled as
\begin{equation}
\label{eq:modelOverall}
    p_{Z|X}(z|x):= \!\!\!\!\!\sum\limits_{y\in \{0, 1\}^{|{\mathcal R}|}} p_{Y|X}(y|x) p_{Z|Y}(z|y)
\end{equation}
The conditional distribution $p_{Z|Y}$ is modelled as 
\begin{eqnarray}
\label{eq:noiseLabel1}
p_{Z|Y}(z|y) & := & \prod\limits_{r\in {\mathcal R}} p_{Z[r]|Y[r]}\left(z[r]|y[r]\right) 
\end{eqnarray}
That is, for each ground-truth label $r\in {\mathcal R}$, $Z[r]$ depends only on $Y[r]$. Furthermore, we assume that for each $r$, there are two parameters $\phi_r^0$ and $\phi_r^1$ governing the dependency of $Z[r]$ on $Y[r]$ via
\begin{eqnarray}
\label{eq:noiseLabel2}
 p_{Z[r]|Y[r]}\!\!\left(\!z[r]|y[r]\right)&\!\!\!\!\!\! := \!\!\!\!\!\!&
 \left\{
 \begin{array}{ll}
\!\!\!\!\phi_r^0 & \!\!\!y[r]\!\!=\!0, \!z[r]\!\!=\!1 \\
\!\!\!\!1\!\!-\!\phi_r^0 & \!\!\!y[r]\!\!=\!0, \!z[r]\!\!=\!0 \\
\!\!\!\!\phi_r^1 & \!\!\!y[r]\!\!=\!1, \!z[r]\!\!=\!0 \\
\!\!\!\!1\!\!-\!\phi_r^1 & \!\!\!y[r]\!\!=\!1, \!z[r]\!\!=\!1 
 \end{array}
 \right.
\end{eqnarray}
We will denote by $\{\phi_r^0, \phi_r^1:r\in {\mathcal R}\}$ collectively by $\phi$. . 

On the other hand, we model $p_{Y|X}$ by
\begin{equation}
\label{forbagmodel}
    p_{Y|X}(y|x):= \prod\limits_{r\in {\mathcal R}} p_{Y[r]|X}\left(y[r]|\overline{x}\right) 
\end{equation}
where $\overline{x}$ is the encoding of bag $x$, implemented using any suitable  neural network. Postponing explaining the details of bag encoding to a later section (namely Section~\ref{bagencodingmodel}), we here specify the form of $p_{Y[r]|X}\left(y[r]|\overline{x}\right)$ for each $r$:
\begin{equation}
\label{eq:groundTruthClassifier}
    p_{Y[r]|X}\left(1|\overline{x}\right) = \sigma \left(\overline{r}^T \overline{x} + b_{r} \right)
\end{equation}
where $\sigma(\cdot)$ is the sigmoid function, $\overline{r}$ is a $|{\mathcal R}|$-dimensional vector and $b_{r}$ is bias. That is, $p_{Y[r]|X}$ is essentially a logistic regression (binary) classifier based on the encoding $\overline{x}$ of bag $x$.  We will denote by $\theta$ the set of all parameters 
$\{\overline{r}: r\in {\mathcal R}\}$ and the parameters for generating encoding $\overline{x}$. 

\begin{figure}[ht!]
	\centering 
	\includegraphics[width=.805\columnwidth]{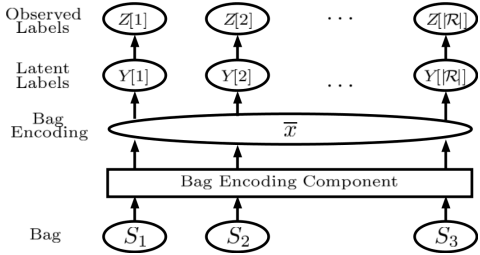}
	\caption{The probabilistic framework of nEM}
	\label{fig:framework}   
\end{figure}

At this end, we have defined the overall structure of the nEM framework (summarized in 
Figure \ref{fig:framework}). Next, we will discuss the learning of it. 
\subsection{Learning with the EM Algorithm}
Let $\ell (z|x; \phi, \theta)$ be the log-likelihood of observing the label set $z$ given the bag $x$, that is, 
\begin{equation}
    \ell (z|x; \phi, \theta): = \log p_{Z|X}(z|x).
\end{equation}
The structure of the framework readily enables a principled learning algorithm based on the EM algorithm \cite{Dempster:77}. 

Let $\ell_{\rm total}(\phi, \theta)$ be the log-likelihood  defined as
\begin{eqnarray}
\label{eq:loss}
    \ell_{\rm total}(\phi, \theta) &:= & \sum_{b \in {\mathcal B}} \ell (z_b|x_b; \phi, \theta).
\end{eqnarray}
The learning problem can then be formulated as maximizing this objective function over its parameters $(\phi, \theta)$, or solving
\begin{equation}
\label{eq:optOriginal}
(\widehat{\phi}, \widehat{\theta}): = \arg \max_{\phi, \theta} \ell_{\rm total}(\phi, \theta).
\end{equation}
Let $Q$ be an arbitrary distribution over $\{0, 1\}^{|{\mathcal R}|}$.
Then it is possible to show

\begin{equation}\small
\begin{split}
&\ell(z|x; \phi, \theta) \\
&\!\!\!= \!
\log \!\!\!\!\!\!\sum\limits_{y\in \{0, 1\}^{|{\mathcal R}|}}
\!\!\!\!\!\!
Q(y)
\frac{
\prod\limits_{r\in {\mathcal R}}
\!\!p_{Y[r]|X}(y[r]|\overline{x})
p_{Z[r]|Y[r]}(z[r]|y[r])
}
{Q(y)} \\
&\!\!\!\geqslant \!\!\!\!
\!\!\!
\sum\limits_{y\in \{0, 1\}^{|{\mathcal R}|}} \!\!\!\!\!\!
\!
Q(y)
\log \!
\frac{
\prod\limits_{r\in {\mathcal R}}
\!\!p_{Y[r]|X}(y[r]|\overline{x})
p_{Z[r]|Y[r]}(z[r]|y[r])
}
{Q(y)} \\
&\!\!\!=
{\mathcal L}(z|x; \phi, \theta, Q)
\end{split}
\end{equation}
where the lower bound ${\mathcal L}(z|x; \phi, \theta, Q)$,  often referred to as the 
variational lower bound. Now we define ${\mathcal L}_{\rm total}$ as
\begin{eqnarray}
    {\mathcal L}_{\rm total}(\!\phi, \!\theta, \!\{Q_{b}\!:\!b\!\in\! {\mathcal B}\}\!) 
    \!\!:= \!\!\!\sum\limits_{b\in {\mathcal B}} \!\!{\mathcal L} \left(z_b|x_b;\! \phi, \!\theta,\! Q_b\right)
\end{eqnarray}
where we have introduced a $Q_b$ for each $b\in {\mathcal B}$.
Instead of solving the original optimization problem (\ref{eq:optOriginal}), we can turn to solving a different optimization problem by maximizing ${\mathcal L}_{\rm total}$
\begin{equation}
\label{eq:optEM}
\begin{split}
\!\!\!(\widehat{\phi}, \widehat{\theta}, \{\widehat{Q}_b\})\!:=\! \arg\!\!\!\max_{\phi, \theta, \{Q_b\}}\!\! {\mathcal L}_{\rm total}(\phi, \theta, \{Q_b\})
\end{split}   
\end{equation}

The EM algorithm for solving the optimization problem (\ref{eq:optEM}) is essentially the coordinate ascent algorithm on objective function ${\mathcal L}_{\rm total}$, where we iterate over two steps, the E-Step and the M-Step.  In the E-Step, we maximizes 
${\mathcal L}_{\rm total}$ over  $\{Q_b\}$ for the current setting of $(\phi, \theta)$ and in the M-Step, we 
maximize ${\mathcal L}_{\rm total}$ $(\phi, \theta)$ for the current setting of $\{Q_b\}$. We now describe the two steps in detail, where we will use superscript $t$ to denote the iteration number. 

\noindent \underline{\bf E-step}: In this step, we hold $(\phi, \theta)$ fixed and update $\{Q_b:b\in {\mathcal B}\}$ to maximize 
the lower bound $\mathcal{L}_{\rm total}(\phi, \theta, \{Q_b\})$. This boils down to update each factor $Q_{b,r}$ of $Q_b$ according to: 
\begin{equation}\small
\label{eq:estep}
\begin{split}
\!\!\!\!Q_{b, r}^{t+1}(y[r]) \!:&\! = \!p_{Y[r]|X,Z}(y_b[r]|x_b,z_b[r];\theta^{t}, \phi_{r}^{t}) \\
& = \!\frac{p_{Z|Y}(z_b[r]|y_b[r]; \phi_{r}^{t}) p_{Y|X}(y_b[r]|x_b; \theta^{t})}{p_{Z|X}(z_b[r]|x_b; \theta^{t}, \phi_{r}^{t})} \\
& = \!\frac{p_{Z|Y}(z_b[r]|y_b[r]; \phi_{r}^{t}) p_{Y|X}(y_b[r]|x_b; \theta^{t})}{\!\!\!\!\!\!\!\!\sum\limits_{y[r]\in \left \{ 0, 1 \right \}}\!\!\!\!\!\!\!p_{Z|Y}(z_b[r]|y[r]; \phi_{r}^{t}) p_{Y|X}(y[r]|x_b; \theta^{t})}			   		   				   
\end{split}
\end{equation}

\noindent \underline{\bf M-step}: In this step, we hold $\{Q_b\}$ fixed and update $(\phi, \theta)$, 
 to maximize the lower bound $\mathcal{L}_{\rm total}(\phi, \theta, \{Q_b\}) $. Let
\begin{equation}\small
\label{eq:mstep}
\begin{split}
M = \prod\limits_{r\in {\mathcal R}}\!\!p_{Y[r]|X}(y_b[r]|\overline{x}_b)p_{Z[r]|Y[r]}(z_b[r]|y_b[r]),
\end{split}
\end{equation}
then $(\phi, \theta)$ is updated according to: 
\begin{equation}\small
\label{eq:mstep}
\begin{split}
	&(\theta^{t+1}, \phi^{t+1}) 
	:= \arg\!\max_{\theta, \phi}\mathcal{L}_{\rm total}(\theta, \phi, \{Q_b^{t+1}\}) \\
	& \!\!\!= \!\arg\!\max_{\theta, \phi}\!\!\!
	\sum\limits_{b\in {\mathcal B}}\!\sum\limits_{y_b\in \{0, 1\}^{|{\mathcal R}|}}\!\!\!\!\!\!\!\!\!\!Q_{b}^{t+1}(y)\!\log \frac{M}{Q_{b}^{t+1}(y)} \\
	& \!\!\!= \!\arg\!\max_{\theta, \phi}\!\!\!
	\sum\limits_{b\in {\mathcal B}}\!\sum\limits_{y_b\!\in\! \{\!0,\! 1\!\}^{\!|\!{\mathcal R}\!|}}\!\!\!\!\!\!\!\!Q_{\!b}^{t\!+\!1}\!(y)\!\log \!M\!\! - \!\!\!\!\sum\limits_{\!\!b\!\in {\mathcal B}}\!\!\sum\limits_{y_b\!\in\!\left \{\! 0,\! 1 \!\right \}^{\!|\!{\mathcal R}\!|}}\!\!\!\!\!\!\!\!Q_{b}^{t}(y)\!\log \!Q_{b}^{t+1}(y) \\
	& \!\!\!= \!\arg\!\max_{\theta, \phi}\!\!\!\sum\limits_{b\in {\mathcal B}} \!\sum\limits_{r\in {\mathcal R}}\!\sum\limits_{y_b\![\!r\!]\in\!\left \{ \!0, \!1\! \right \}}\!\!\!\!\!\!\!Q_{\!b,r}^{t+1}\!(y[r])\!\left (\log p_{Z|Y}(z_b[r]|y_b[r];\phi_{r})\right. \\
	&\phantom{=\;\;} \left. \! + \log p_{Y|X}(y_b[r]|\overline{x}_b; \theta)\right )						    
\end{split}
\end{equation}
Overall the EM algorithm starts with initializing $(\phi, \theta, \{Q_b\})$, and then iterates over the two steps until convergence or after some prescribed number of iterations is reached. There are however several caveats on which caution is needed. 
First, the optimization problem in the M-Step cannot be solved in closed form. As such we take a stochastic gradient descent (SGD) approach\footnote{In fact, the approach should be called ``stochastic gradient ascent'', since we are maximizing the objective function not minimizing it.
}. 

In each M-Step, we perform $\Delta$ updates, where $\Delta$ is a hyper-parameter.  Such an approach is sometimes referred to as ``Generalized EM'' ~\cite{Wu:83,Jojic:01,Greff:17}. Note that since the parameters $\theta$ are parameters for the neural network performing bag encoding, the objective function ${\mathcal L}_{\rm total}$ is highly non-convex with respect to $\theta$. This makes it desirable to choose an appropriate $\Delta$. Too small $\Delta$ results in little change in $\theta$ and hence provides insufficient signal to update $\{Q_b\}$; too large $\Delta$, particularly at the early iterations when $\{Q_b\}$ has not been sufficiently optimized, tends to make the optimization stuck in undesired local optimum. In practice, one can try several values of $\Delta$ by inspecting their achieved value of ${\mathcal L}_{\rm total}$ and select the $\Delta$ giving rise to the highest 
${\mathcal L}_{\rm total}$. Note that such a tuning of $\Delta$ requires no access of the testing data.
The second issue is that the EM algorithm is known to be sensitive to initialization. In our implementation, in order to provide a good initial parameter setting, we set each $Q_b^0$ to $z_b$. Despite the fact that $z$ contains noise, this is a much better approximation of the posterior of true labels than any random initialization. 

\input{bagEncoding.tex}

%% file: bagEncoding.tex
\begin{figure}[h]
	\centering 
	\includegraphics[width=0.79\columnwidth]{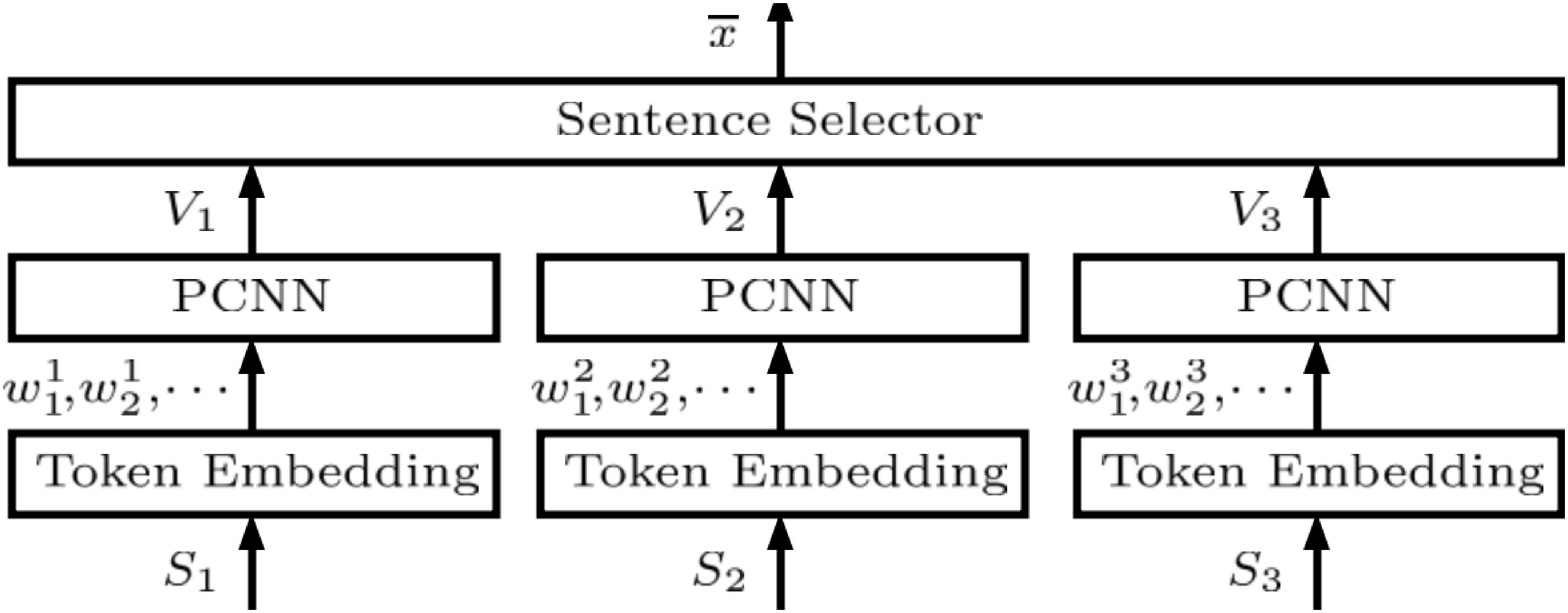}
	\caption{The structure of Bag Encoding component.}
	\label{fig:bagEncoding}	
\end{figure}
The nEM framework needs the encoding of a bag of $x$ as discussed in Equation~\ref{forbagmodel}. Any  suitable neural network can  be deployed to achieve this goal. Next, we present the widely used methods for DS relation extraction strategies: the Bag Encoding Models. 
\subsection{Bag Encoding Models}
\label{bagencodingmodel}

As illustrated in Figure~\ref{fig:bagEncoding}, the Bag Encoding Models include three components:  Word-Position Embedding, Sentence Encoding, and Sentence Selectors. 
\subsubsection{Word-Position Embedding}
For the $j^{\rm th}$ word in a sentence, Word-Position Embedding generates a vector representation ${\bf w}_j$ as  concatenated three components $\left [ {\bf w}_j^w, {\bf w}_j^{p1}, {\bf w}_j^{p2} \right ]$. Specifically, $ {\bf w}_j^w$ is the word embedding of the word and ${\bf w}_j^{p1}$ and ${\bf w}_j^{p2}$ are two position embeddings. Here ${\bf w}_j^{p1}$ (resp. ${\bf w}_j^{p2}$) are the embedding of the 
relative location of the word with respect to the first (resp. second) entity in the sentence. The dimensions of word and position embeddings are denoted by $d_w$ and $d_p$ respectively. 
\subsubsection{Sentence Encoding}
Sentence encoding uses Piecewise Convolutional Neural Networks (PCNN)~\cite{Zeng:15, Lin:16, Ye:17, Ji:17}, which consists of convolution followed by Piecewise Max-pooling.
The convolution operation uses a list of matrix kernels such as $ \left \{{\bf K}_{1}, {\bf K}_{2}, \cdots, {\bf K}_{l_{ker}}\right \} $ to extract {\em n-gram} features through sliding windows of length $ l_{win} $. Here $ {\bf K}_{i}\in \mathbb{R}^{l_{win}\times d_{e}} $ and $ l_{ker} $ is the number of kernels. Let $ {\bf w}_{j-l_{win}+1:j} \in \mathbb{R}^{l_{win} \times d_{e}} $ be the concatenated vector of token embeddings in the $ j^{\rm th} $ window. The output of convolution operation is a matrix $ {\bf U}\in \mathbb{R}^{l_{ker}\times (m+l_{win}-1)} $ where each element is computed by
\begin{equation}\label{eq:conv}
{\bf U}_{ij} = {\bf K}_{i}\odot{\bf w}_{j-l_{win}+1:j}+b_{i}
\end{equation}
where $\odot$ denotes inner product. 
In Piecewise Max-Pooling, ${\bf U}_{i} $ is segmented to three parts $ \left \{{\bf U}_{i}^{1}, {\bf U}_{i}^{2}, {\bf U}_{i}^{3} \right \} $ depending on whether an element in ${\bf U}_i$ is on the left or right of the two entities, or between the two entities. Then max-pooling is applied to each segment, giving rise to
\begin{equation}\label{eq:maxpool}
{\bf g}^{ip} = \max({\bf U}_{i}^{p}), 1\leqslant i \leqslant l_{ker}, 1\leqslant p \leqslant 3
\end{equation}
Let $ {\bf g}=\left [{\bf g}^{1}, {\bf g}^{2}, {\bf g}^{3} \right ] $.
Then sentence encoding outputs 
\begin{equation}\label{eq:v}
{\bf v} = {\rm ReLU} ({\bf g})\circ {\bf h}, 
\end{equation}
where $ \circ $ is element-wise multiplication and $ {\bf h} $ is a vector of Bernoulli random variables, representing dropouts.  
\subsubsection{Sentence Selectors}
Let $n$ be the number of sentences in a bag. We denote a matrix ${\bf V}\in \mathbb{R}^{n\times (l_{ker}\times3)} $consisting of each sentence vector $ {\bf v}_{T} $. $ {\bf V}_{k:} $ and $ {\bf V}_{:j} $ are used to index the $k^{\rm th}$ row vector and $j^{\rm th}$ column vector of $ {\bf V} $ respectively. Three kinds of sentence-selectors are used to construct the bag encoding.
\noindent \underline{Mean-selector~\cite{Lin:16, Ye:17}}: The bag encoding is computed as
\begin{equation}\label{eq:B0}
\overline{x} = \frac{1}{n}\sum_{k=1}^{n}{\bf V}_{k:}
\end{equation}
\noindent \underline{Max-selector~\cite{Jiang:16}}: 
The $j^{\rm th}$ element of bag encoding $\overline{x}$ is computed as
\begin{equation}\label{eq:B1}
\overline{x}_{j} = \max ({\bf V}_{:j})
\end{equation}
\noindent \underline{Attention-selector}
Attention mechanism is extensively used for sentence selection in relation extraction by weighted summing of the sentence vectors, such as in ~\cite{Lin:16, Ye:17,  Su:18}. However, all these works assume that the labels are {\em correct} and only use the golden label embeddings to select sentences at training stage. We instead selecting sentences using all label embeddings $\overline{r}$ and construct a bag encoding for each label $r\in {\mathcal R}$. The output is then a list of vectors $ \left \{\overline{x}^{r} \right \} $, in which the $r^{\rm th}$ vector is calculated through attention mechanism as
\begin{equation}\label{eq:attention}
e_{l} \!\!=\! \!{\bf V}_{\!l:}^{\!T}\!{\bf A}\overline{r}, 	   				  
~\alpha _{k} \!\!=\!\! \frac{\exp(e_{k})}{\sum_{l=\!1}^{n}\!\exp(e_{l})}, 
~\overline{x}^{r}	\!\!\!= \!\!\!\sum_{k=1}^{n}\!\!\alpha _{k}\!\!{\bf V}_{\!\!k:}
\end{equation}
where 
$ {\bf A}\in \mathbb{R}^{d_r \times d_{r}} $ is a diagonal matrix and $d_r$ is the dimension of relation embedding.

%% file: exp.tex
\section{Experimental Study}
We first conduct experiments on the widely used benchmark   data set Riedel~\cite{Riedel:10}, and then on the TARCED~\cite{Zhang:17} data set. The latter allows us to  control the noise level in the labels to observe the behavior and working mechanism of our proposed method. The code for our model is found on the Github page~\footnote{\url{https://github.com/AlbertChen1991/nEM}}.
\input{expds}
\input{expsuper}

%% file: expds.tex
\subsection{Evaluation on the Riedel Dataset}
\label{sec:DS_data}
The Riedel dataset\footnote[2]{\url{http://iesl.cs.umass.edu/riedel/ecml/}} is a widely used DS dataset for relation extraction. It was developed in ~\cite{Riedel:10} through aligning entity pairs from Freebase with the New York Times (NYT) corpus.
There are 53 relations in the Riedel dataset, including ``{\em NA}''. The bags collected from the 2005-2006 corpus are used as the training set, and the bags from the 2007 corpus are used as the test set. The training data contains $281,270$ entity pairs and $522,611$ sentences; the testing data contains $96,678$ entity pairs and $172,448$ sentences.
\input{annot}
\subsubsection{\bf Evaluation Metrics and Baselines}
For the Riedel dataset, we train the models on the noisy training set and test the models on the manually labeled test set. The precision-recall (PR) curve is reported to compare performance of models.
Three baselines are considered: 
\begin{itemize}[leftmargin=*, itemsep=-0.1cm]
	\item PCNN+MEAN~\cite{Lin:16, Ye:17}: A model using PCNN to encode sentences and a mean-selector to generate bag encoding.
	\item PCNN+MAX~\cite{Jiang:16}: A model using PCNN to encode sentences and a max-selector to generate bag encoding.	
	\item PCNN+ATT~\cite{Lin:16, Ye:17,  Su:18}: A model using PCNN to encode sentences and an attention-selector to generate bag encoding.
\end{itemize}
We compare the three baselines with their nEM versions (namely using them as the Bag Encoding component in nEM), which are denoted with a ``+nEM'' suffix.

\subsubsection{\bf Implementation Detail} 
\label{dsprama}
Following previous work, we tune our models using three-fold validation on the training set. As the best configurations, we set $d_{w}=50$, $d_{p}=5$ and $d_{r}=230$. For PCNN, we set $l_{win}=3$, $l_{ker}=230$ and set the probability of dropout to $0.5$. We use Adadelta~\cite{Zeiler:12} with default setting ($\rho = 0.95, \varepsilon = 1e^{-6}$) to optimize the models and the initial learning rate is set as $ 1 $. The batch size is fixed to $160$ and the max length of a sentence is set as $120$. 

For the noise model $p_{Z\mid Y}$, we set $\phi_{NA}^{0}=0.3, \phi_{NA}^{1}=0$ for the NA label and $\phi_{r}^{0}=0, \phi_{r}^{1}=0.9$ for other labels $r\neq NA$. In addition, the number $\Delta$ of SGD updates in M-step is set to $2000$.  

\subsubsection{\bf Predictive performance}
\begin{figure}[h]
	\centering 
	\includegraphics[width=.68\columnwidth]{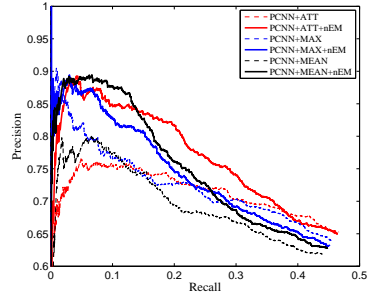}
	\caption{PR curves on DS test set.}
	\label{fig:origin}	
\end{figure}

The evaluation results on manually labeled test set are shown in Figure ~\ref{fig:origin}.
From which, we observe that the PR-curves of the nEM models are above their corresponding baselines by significant margins, especially in the low-recall regime. This observation demonstrates the effectiveness of the nEM model on improving the extracting performance upon the baseline models.
We also observe that models with attention-selector achieve better performance than models with mean-selector and max-selector, demonstrating the superiority of the attention mechanism on sentence selection. 

\begin{figure}[h]
\begin{tabular}{c c}
\hspace{-.5cm}
			\includegraphics[width=.5\columnwidth]{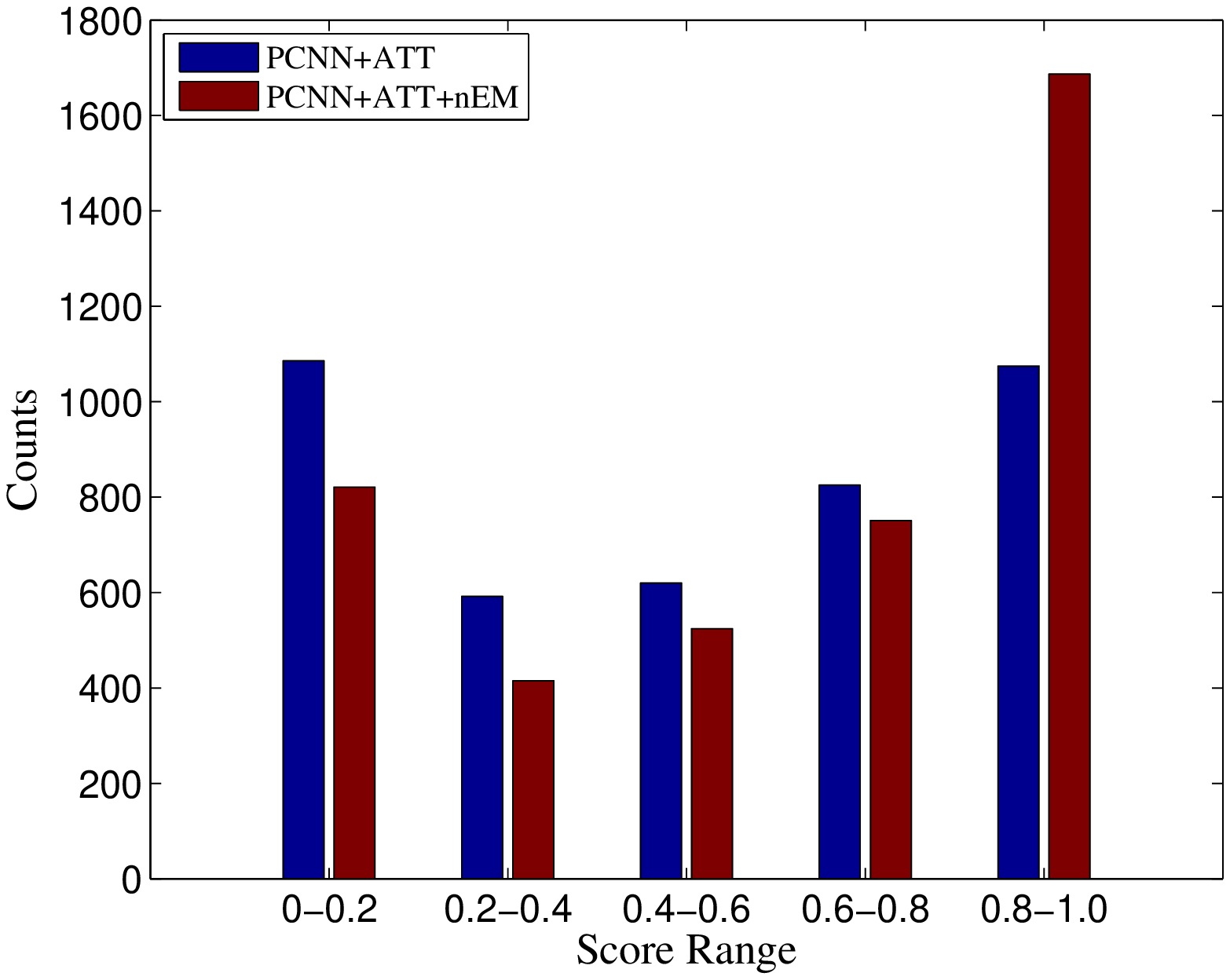}
			&
			\hspace{-.5cm}
			\includegraphics[width=.5\columnwidth]{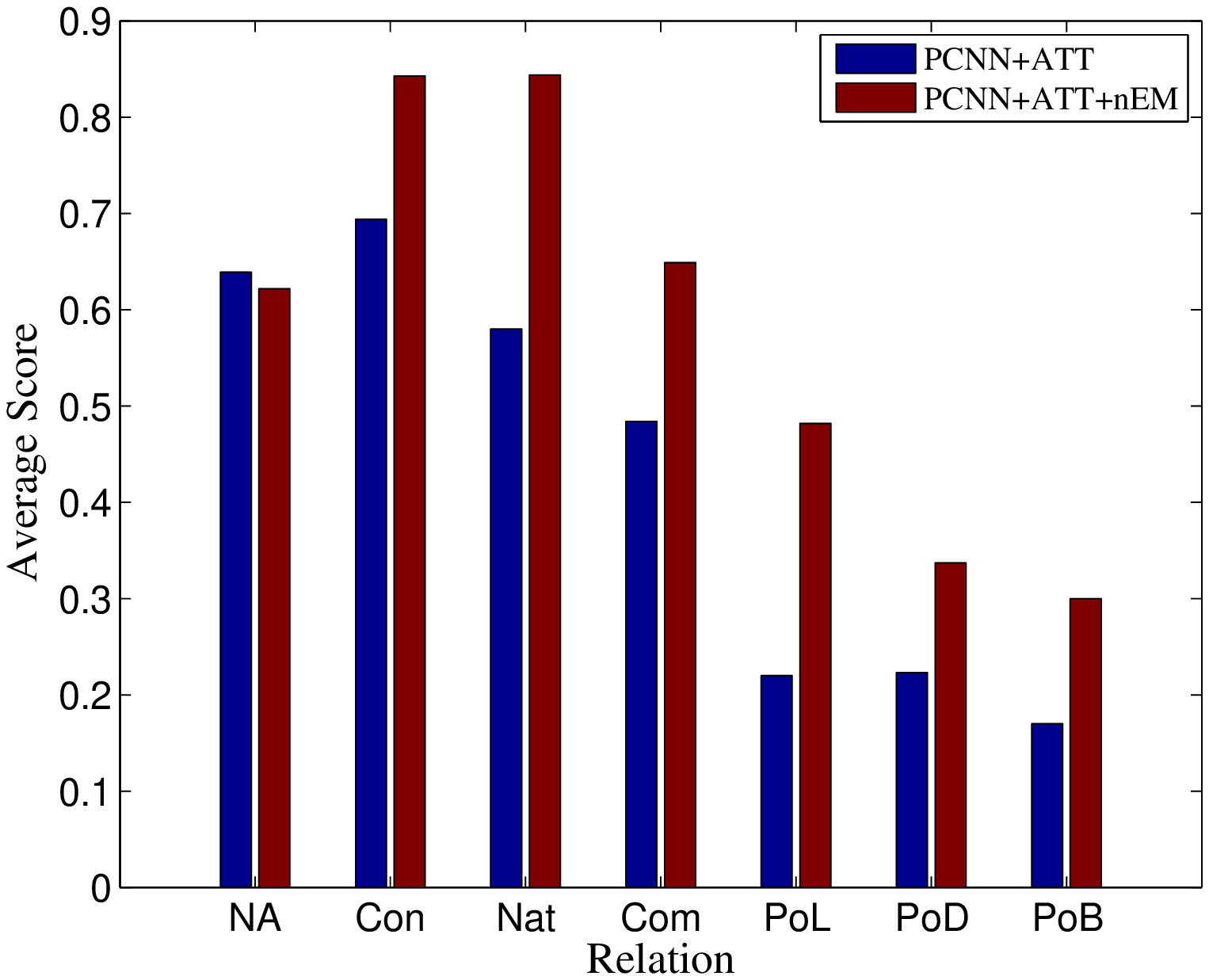}
			\\
			\hspace{-.4cm}\small (a) Score distribution
			&
			\hspace{-.4cm}\small (b) Scores on main relations
\end{tabular}
	\caption{Predictive probabilities on the test set. CON, Nat, Com, PoL, PoD and PoB represent 'contains', 'nationality', 'company', 'place lived', 'place of death' and 'place of birth' relations respectively.\label{fig:count}}
\vspace{-.3cm}
\end{figure}

We then analyze the predicting probability of PCNN+ATT and PCNN+ATT+nEM on the ground-truth labels in the test set. We divide the predicting probability values into $5$ bins and count the number of label within each bin. The result is shown in Figure ~\ref{fig:count}(a). We observe that the count for PCNN+ATT in bins $0.0-0.2$, $0.2-0.4$, $0.4-0.6$ and $0.6-0.8$ are all greater than PCNN+ATT+nEM. But in bin $0.8-1.0$, the count for PCNN+ATT+nEM is about 55\% larger than PCNN+ATT. This observation indicates that nEM can promote the overall predicting scores of ground-truth labels.

Figure ~\ref{fig:count}(b) compares PCNN+ATT and PCNN+ATT+nEM in their predictive probabilities on the frequent relations. The result shows that PCNN+ATT+nEM achieves higher average predicting probability on all these relations, except for the NA relation, on which PCNN+ATT+nEM nearly levels with PCNN+ATT. This phenomenon demonstrates that nEM tends to be more confident in predicting the correct labels. In this case, raising the predictive probability on the correct label does increase the model’s ability to make the correct decision, thereby improving performance.

%% file: annot.tex
\subsubsection{Ground-Truth Annotation}
\label{annotation}

The Riedel dataset contains no ground-truth labels. The held-out evaluation~\cite{Mintz:09} 
is highly unreliable in measuring a model's performance against the ground truth. 
Since this study is concerned with discovering the ground-truth labels, the conventional held-out evaluation is no longer appropriate. For this reason, we annotate a subset of the original test data for evaluation purpose. Specifically, we annotate a bag $x_b$ by its all correct labels, and if no such labels exist, we label the bag NA.

In total 3762 bags are annotated, which include all bags originally labelled as non-NA ("Part 1") and 2000 bags ("Part 2") selected from the bags originally labelled as NA but have relatively low score for NA label under a pre-trained PCNN+ATT~\cite{Lin:16} model. The rationale here is that we are interested in inspecting the model's performance in detecting the labels of known relations rather than NA relation.

\begin{table}[h]\small 
	\centering
	\caption{The statistics of ground-truth annotation. {\em origin} denotes the total number of originally assigned labels of these bags. {\em correct} and {\em wrong} denote the total number of correctly assigned and wrongly assigned labels. {\em added} denote the number of missing labels we added into these bags.}
	\begin{tabular}{|c|c|c|c|c|c|}
		\hline
		&bags&origin&correct&wrong&added\\
		\hline
		Part 1&1762&1953&1245&708&804\\	
		\hline	
		Part 2&2000&2000&938&1062&1211\\	
		\hline											
	\end{tabular}
	\label{tab:manual}
\end{table} 

The statistics of annotation is shown in Table ~\ref{tab:manual}. Through the annotation, we notice that, about 36\% of the original labels in original non-NA bags and 53\% of labels in original NA bags are wrongly assigned. Similar statistics has been reported in previous works ~\cite{Riedel:10, Feng:18}. 

%% file: expsuper.tex
\subsection {Evaluation on the TACRED Dataset}
TACRED is a large supervised relation extraction dataset collected in ~\cite{Zhang:17}. The text corpus is generated from the TAC KBP evaluation of years $2009-2015$. 
Corpus from years $2009-2012$, year $2013$ and year $2014$ are used as training, validation and testing sets respectively. In all annotated sentences, $68,124$ are used for training, $22,631$ for validation and $15,509 $ for testing. There are in total $42,394$ {\it distinct} entity pairs, where $33,079$ of these entity pairs appear in exactly one sentence. Since these $33,079$ entity pairs dominate the training set, we simply treat each sentence as a bag for training and testing. Another advantage of constructing single-sentence bags is that it allows us to pinpoint the correspondence between the correct label and its supporting sentence. The number of relations in TACRED dataset is $42$, including a special relation ``{\em no relation}'', which is treated as NA in this study.

\subsubsection{Constructing Semi-synthetic Dataset}
To obtain insight into the working of nEM, we create a simulated DS dataset by inserting noisy labels into the training set of a supervised dataset, TACRED. Since the training set of TACRED was manually annotated, the labels therein may be regarded as the ground-truth labels. Training using this semi-synthetic dataset allows us to easily observe models' behaviour with respect to the noisy labels and the true labels.

We inject artificial noise into the TACRED training data through the following precedure. For each bag $ x_b $ in the training set, we generate a noisy label vector $\tilde{z}_b$ from the observed ground-truth label vector $y_b$. Specifically, $\tilde{z}_b$ is generated by flipping each element $y_b[r]$ from $ 0 $ to $ 1 $ or $ 1 $ to $ 0 $ with a probability $ p_{f}$. This precedure simulates a DS dataset through introducing wrong labels into training bags, thus corrupts the training dataset.

\subsubsection{\bf Experimental Settings}

Following ~\cite{Zhang:17}, the common relation classification metrics Precision, Recall and F1 are used for evaluation. 
The PCNN model is used to generate the bag encoding since sentence selection is not needed in this setting. The same hyper-parameter settings as in Section ~\ref{dsprama} are used in this experiment. For the noise model $p_{Z[r]|Y[r]}$ of PCNN +nEM, we set $\phi_{r}^{0}=0.1, \phi_{r}^{1}=0.1$ for each label $ r \in {\mathcal R}$. The number $\Delta$ of SGD updates in each M-step is set to $1600$. 
\subsubsection{\bf Test Results}
From Table ~\ref{tab:TACRED}, we see that PCNN+nEM achieves better recall and F1 score than the PCNN model under various noise levels.  Additionally, the recall and F1 margins between PCNN and PCNN+nEM increase with noise levels. This suggests that nEM keeps better performance than the corresponding baseline model under various level of training noise. We also observe that the precision of nEM is consistently lower than that of PCNN when noise is injected to TACRED. This is a necessary trade-off present in nEM. The training of nEM regards the training labels with less confidence based on noisy-label assumption. This effectively lowers the probability of seen training labels and considers the unseen labels also having some probability to occur. When trained this way, nEM, at the prediction time, tends to predict more labels than its baseline (PCNN) does.  Note that in TACRED, each instance contains only a single ground truth label. Thus the tendency of nEM to predict more labels leads to the reduced precision. However, despite this tendency, nEM, comparing with PCNN, has a stronger capability in detecting the correct label and gives the better recall of nEM. The gain in recall dominates the loss in precision.

\begin{table}[h]
	\centering
	\caption{Test performance on the TACRED dataset.}	
	\scalebox{0.999}{
	\begin{tabular}{|c|c|c|c|c|c|c|}
		\hline
		\multirow{2}*{$p_{f}$}&\multicolumn{3}{c|}{PCNN}&\multicolumn{3}{c|}{PCNN+nEM}\\
		\cline{2-7}&\textbf{P}&\textbf{R}&\textbf{F1}&\textbf{P}&\textbf{R}&\textbf{F1}\\		
		\hline
		0.02&65.2&30.0&41.1&61.8&34.8&44.5\\	
		\hline
		0.04&63.5&28.9&39.7&60.2&33.8&43.3\\
		\hline
		0.06&70.5&25.2&37.1&59.9&31.1&41.0\\
		\hline
		0.08&63.5&23.6&34.4&58.8&29.8&39.6\\
		\hline	
		0.10&66.1&22.7&33.9&56.5&28.9&38.3\\
		\hline									
	\end{tabular}}
	\label{tab:TACRED}
\end{table}

\subsubsection{\bf Training Label Probabilities}
\begin{figure}[h]
	\begin{tabular}{cc}
		\hspace{-.4cm}
		\includegraphics[width=.5\columnwidth]{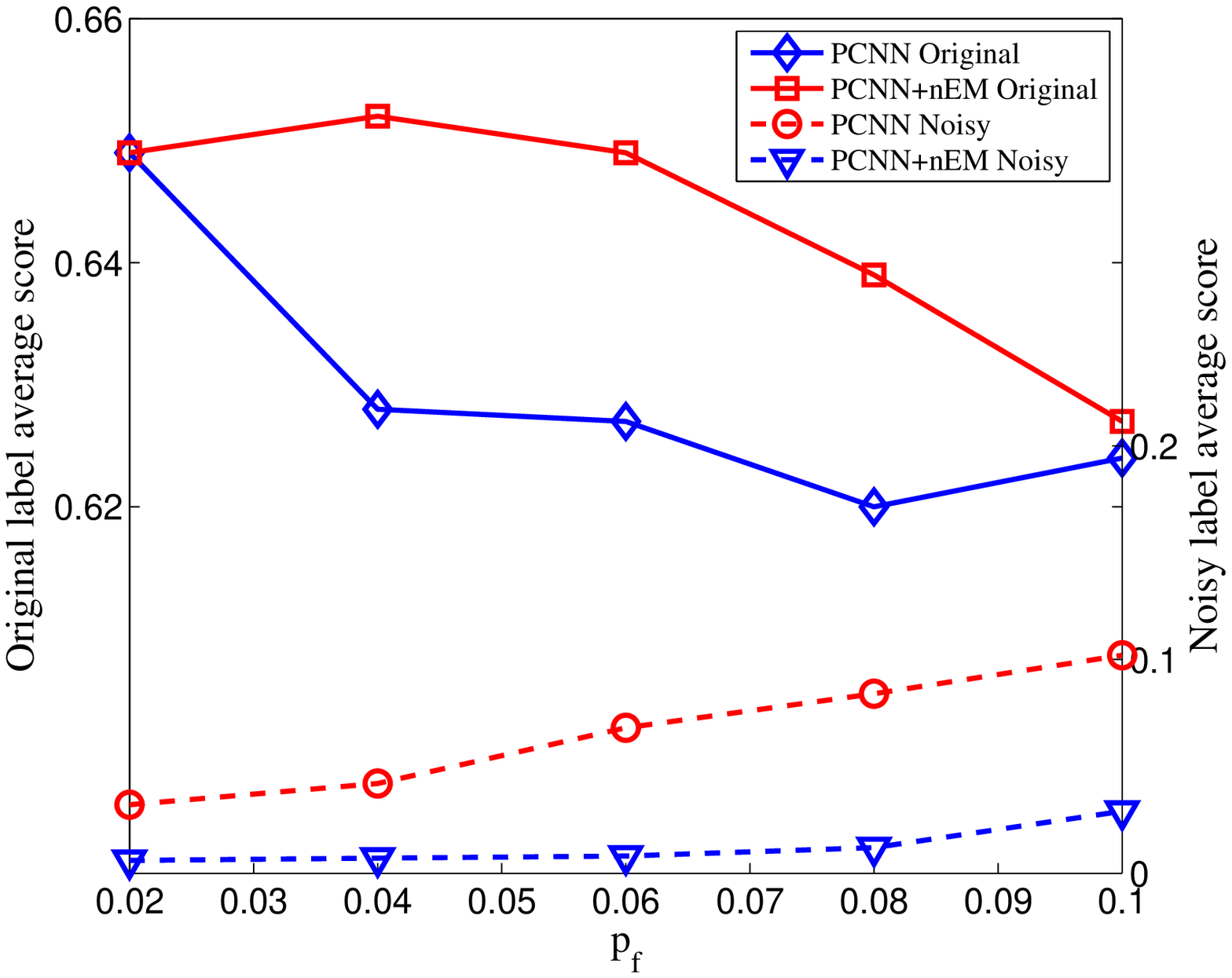}
		\hspace{-.4cm}
		& 
		\includegraphics[width=.5\columnwidth]{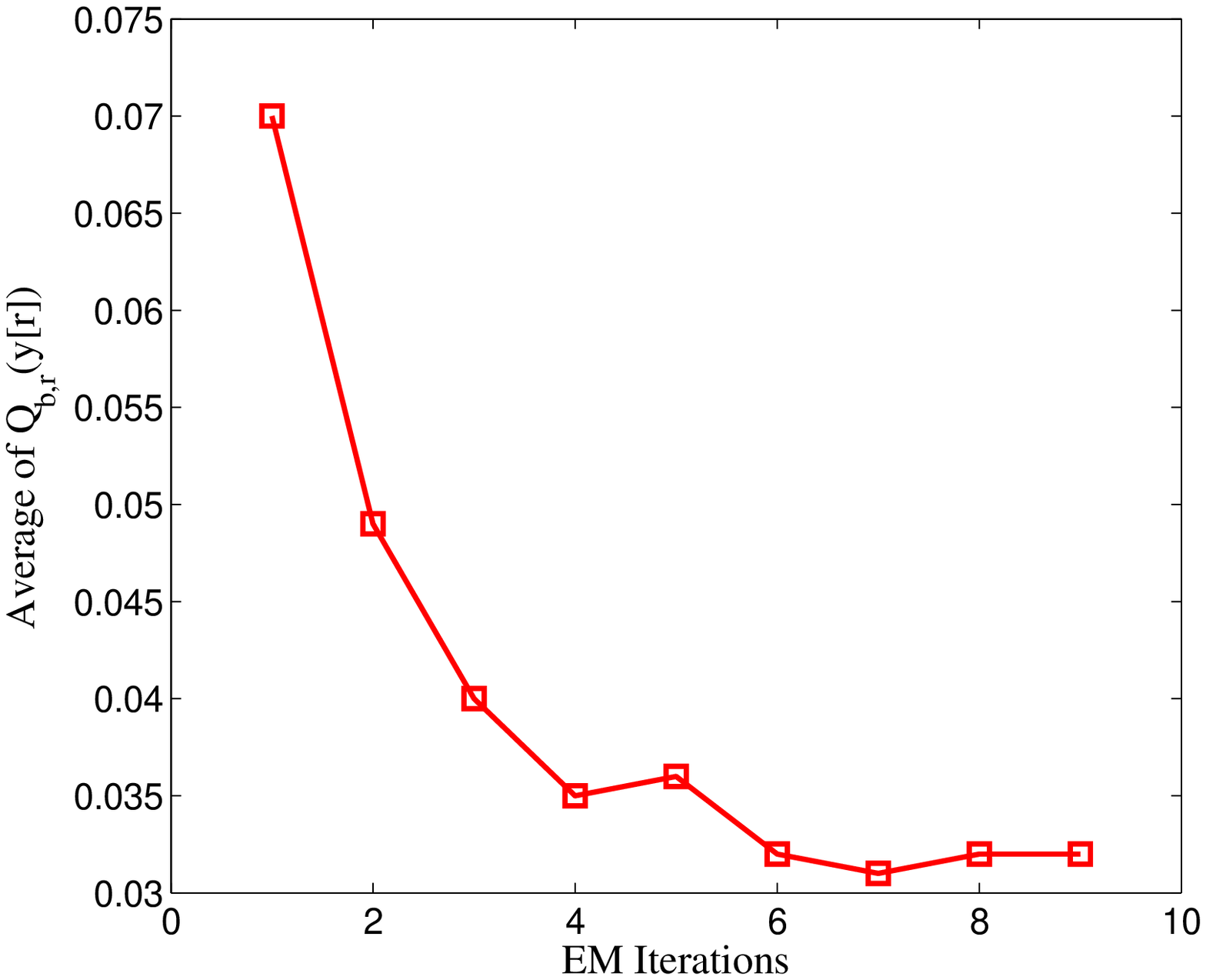}
	\end{tabular}	
	\vspace{-0.1cm}	
	\caption{Average probabilities for noisy and original labels (left) and average $Q_{b,r}(y[r])$ over EM iterations (right)}
	\label{fig:ave_train}
	\vspace{-0.3cm}
\end{figure}

The predicting probabilities for the noise labels and the original true labels are also evaluated under the trained models. The results are shown in Figure ~\ref{fig:ave_train} (left) which reveals that with increasing noise, the average probability for noisy label sets of PCNN and PCNN+nEM both increase and average scores for original label sets of PCNN and PCNN+nEM both decrease. 
The performance degradation of PCNN and PCNN+nEM under noise appears different. The average probability for noisy label sets rises with a higher slope in PCNN than in PCNN+nEM. 
Additionally, the average probability for the original label sets of PCNN+nEM is higher than or equal to PCNN at all noise levels. These observations  confirms that the denoising capability of nEM is learned from effectively denoising the training set.
\subsubsection{\bf Effectiveness of EM Iterations}
For each training bag $x_b$ and each artificial noisy label $r$, we track the probability $Q_{b,r}(y[r])=1$ over EM iterations. This probability, measuring the likelihood of the noise label $r$ being correct, is then averaged over $r$ and over all bags $x_b$. It can be seen in Figure ~\ref{fig:ave_train} (right) that the average value of $Q_{b,r}(y[r])$ decreases as the training progresses, leading the model to gradually ignore noisy labels. This demonstrates the effectiveness of EM iterations and validates the proposed EM-based framework.

%% file: conclusion.tex
\section{Concluding Remarks}
We proposed a nEM framework to deal with the noisy-label problem in distance supervision relation extraction. We empirically demonstrated the effectiveness of the nEM framework, and provided insights on its working and behaviours through data  with controllable noise levels.
Our framework is a combination of latent variable models in probabilistic modelling with contemporary deep neural networks. Consequently, it naturally supports a training algorithm which elegantly nests the SGD training of any appropriate neural network inside an EM algorithm.  
We hope that our approach and the annotated clean testing data would inspire further research along this direction. 